\documentclass[10pt,twocolumn,letterpaper]{article}

\usepackage{wacv}
\usepackage{times}
\usepackage{epsfig}
\usepackage{graphicx}
\usepackage{amsmath}
\usepackage{amssymb}

\usepackage{algorithm}
\usepackage{algorithmic}
\usepackage{multirow}
\usepackage{array}

\def\wacvPaperID{778} 

\wacvfinalcopy 

\ifwacvfinal
\def\assignedStartPage{9876} 
\fi

\ifwacvfinal
\usepackage[breaklinks=true,bookmarks=false]{hyperref}
\else
\usepackage[pagebackref=true,breaklinks=true,colorlinks,bookmarks=false]{hyperref}
\fi

\ifwacvfinal
\else
\fi

\begin{document}

\title{Federated Face Presentation Attack Detection}

\author{Rui Shao\textsuperscript{1}  \qquad Pramuditha Perera\textsuperscript{2} \qquad Pong C. Yuen\textsuperscript{1} \qquad Vishal M. Patel\textsuperscript{2} \\
	\textsuperscript{1}Department of Computer Science, Hong Kong Baptist University\\
	\textsuperscript{2}Department of Electrical and Computer Engineering, Johns Hopkins University
}

\maketitle

\begin{abstract}
Face presentation attack detection (fPAD) plays a critical role in the modern face recognition pipeline. A face presentation attack detection model with good generalization can be obtained when it is trained with face images from different input distributions and different types of spoof attacks. In reality, training data (both real face images and spoof images) are not directly shared between data owners due to legal and privacy issues. In this paper, with the motivation of circumventing this challenge, we propose Federated Face Presentation Attack Detection (FedPAD) framework. FedPAD simultaneously takes advantage of rich fPAD information available at different data owners while preserving data privacy. In the proposed framework, each data owner (referred to as \textit{data centers}) locally trains its own fPAD model. A server learns a global fPAD model by iteratively aggregating model updates from all data centers without accessing private data in each of them. Once the learned global model converges, it is used for fPAD inference. We introduce the experimental setting to evaluate the proposed FedPAD framework and carry out extensive experiments to provide various insights about federated learning for fPAD.
\end{abstract}

\section{Introduction}

\begin{figure}[!htb]
	
	\begin{center}
		
		\includegraphics[ width=0.8\linewidth]{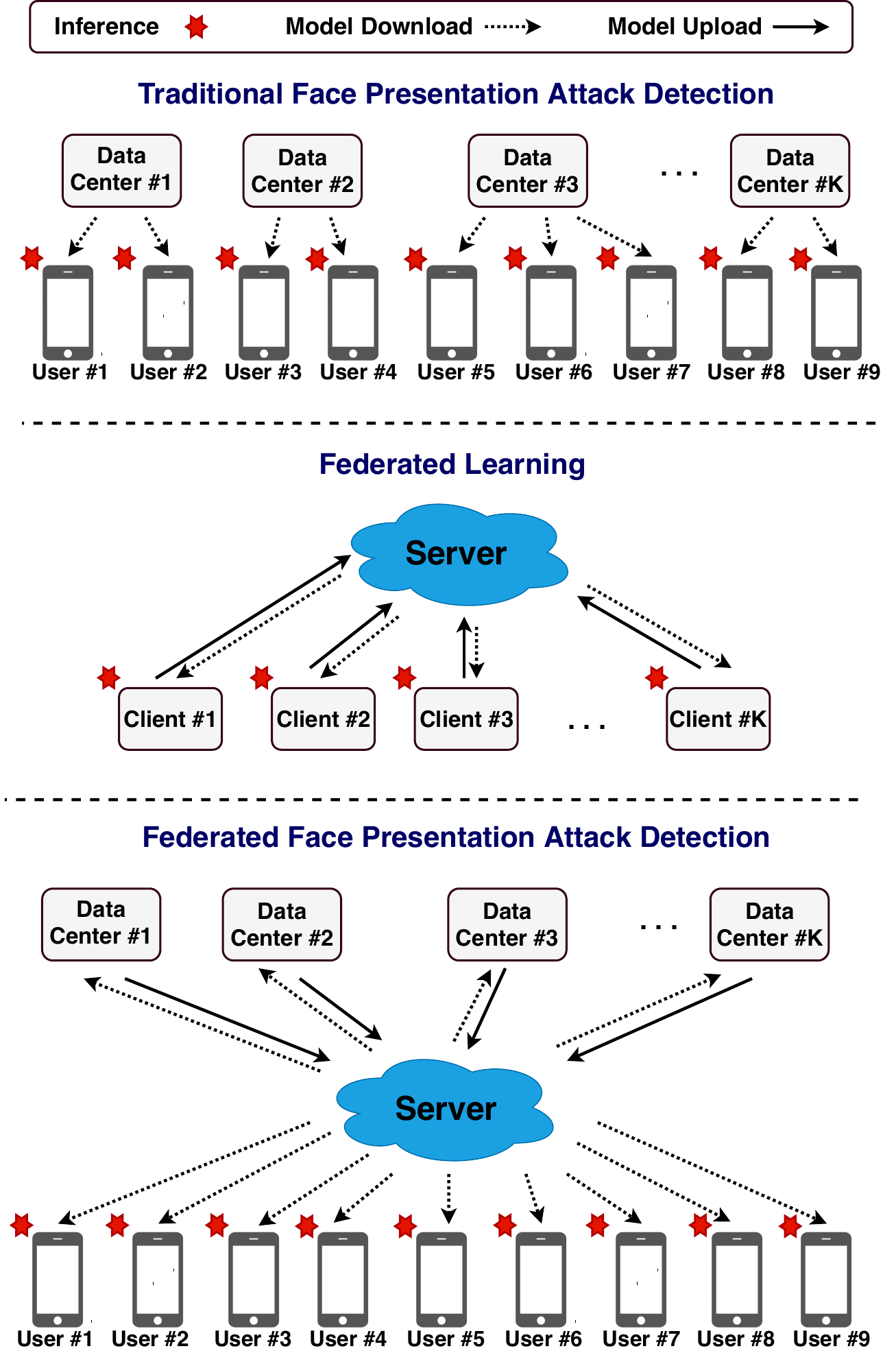}
		
	\end{center}
	
	\caption{Comparison between fPAD (top), traditional federated learning (middle) and the proposed FedPAD (bottom). FedPAD can be a regarded as a special case of traditional federated learning.}
	\label{fig:illustration}
\end{figure}

Recent advances in face recognition methods have prompted many real-world applications, such as automated teller machines (ATMs), mobile devices, and entrance guard systems, to deploy this technique as an authentication method. Wide usage of this technology is due to both high accuracy and convenience it provides. However,  many recent works~\cite{2011IJCBmstexture,2016TIFScolortxt,2015TIFSida,RuiShao2018IJCB,2018CVPRauxliary,Shao2019CVPR,2018TIFSdynamictext,Shao_2020_AAAI} have found that this technique is vulnerable to various face presentation attacks such as print attacks, video-replay attacks~\cite{2017FGoulu,2012ICBcasia,2012BIOSIGidiap,2015TIFSida,2018CVPRauxliary} and 3D mask attacks~\cite{2018ECCVrPPG,2016ECCVrPPG}. Therefore, developing face presentation attack detection (fPAD) methods that make current face recognition systems robust to face presentation attacks has become a topic of interest in the biometrics community.

In this paper, we consider the deployment of a fPAD system in a real-world scenario. We identify two types of stakeholders in this scenario -- \textit{data centers} and \textit{users}. \textit{Data centers} are entities that design and collect fPAD datasets and propose fPAD solutions. Typically \textit{data centers} include research institutions and companies that carry out the research and development of fPAD. These entities have access to both \textit{real data} and \textit{spoof data} and therefore are able to train fPAD models. Different \textit{data centers} may contain images of different identities and different types of \textit{spoof data}. However, each \textit{data center} has  limited data availability. Real face images are obtained from a small set of identities and spoof attacks are likely to be from  a few known types of attacks. Therefore, these fPAD models have poor generalization ability~\cite{Shao2019CVPR,Shao_2020_AAAI} and are likely to be vulnerable against attacks unseen during training.

On the other hand, \textit{users} are individuals or entities that make use of fPAD solutions. For example, when a fPAD algorithm is introduced in mobile devices, mobile device customers are identified as \textit{users} of the fPAD system.  \textit{Users} have access only to \textit{real data} collected from local devices. Due to the absence of \textit{spoof data}, they cannot locally train fPAD models. Therefore, each \textit{user} relies on a model developed by a \textit{data center} for fPAD as shown in Figure~\ref{fig:illustration} (top). Since \textit{data center} models lack generalization ability, inferencing with these models are likely to result in erroneous predictions. 

It has been shown that utilizing \textit{real data} from different input distributions and \textit{spoof data} from different types of spoof attacks through domain generalization and meta-learning techniques can significantly improve the generalization ability of fPAD models~\cite{Shao2019CVPR,Shao_2020_AAAI}. Therefore, the performance of fPAD models, shown in Figure~\ref{fig:illustration} (top), can be improved if data from all \textit{data centers} can be exploited collaboratively. In reality, due to data sharing agreements and privacy policies, \textit{data centers} are not allowed to share collected fPAD data with each other. For example, when a \textit{data center} collects face images from individuals using a social media platform, it is agreed not to share collected data with third parties. 

In this paper, we present a framework called Federated Face Presentation Attack Detection (FedPAD) based on the principles of Federated Learning (FL) targeting fPAD. The proposed method exploits information across all \textit{data centers} while preserving data privacy. In the proposed framework, models trained at \textit{data centers} are shared and aggregated while training images are kept private in their respective \textit{data centers}, thereby preserving privacy.  

Federate learning is a distributed and privacy preserving machine learning technique~\cite{mcmahan2016communication,li2019federated,smith2017federated,sahu2018convergence,mohri2019agnostic}. FL training paradigm defines two types of roles named \textit{server} and \textit{client}.  \textit{Clients} contain training data and the capacity to train a model. As shown in Fig.~\ref{fig:illustration} (middle), each client trains its own model locally and uploads them to the \textit{server} at the end of each training iteration. \textit{Server} aggregates local updates and produces a global model. This global model is then shared with all clients which will be used in their subsequent training iteration. This process is continued until the global model is converged. During the training process, data of each client is kept private. Collaborative FL training allows the global model to exploit rich local clients information while preserving data privacy.   

In the context of FedPAD, both \textit{data centers} and \textit{users} can be identified as clients. However, roles of \textit{data centers}  and \textit{users} are different from conventional clients found in FL. In FL, all \textit{clients} train models and carry out inference locally. In contrast, in FedPAD, only \textit{data centers} carry out local model training. \textit{Data centers} share their models with the \textit{server} and download the global model during training. On the other hand, \textit{users} download the global model at the end of the training procedure and only carry out inference as shown in Figure~\ref{fig:illustration} (bottom).

Main contributions of our paper can be summarized as follows:

\noindent 1. This paper is the first to study the federated learning technique for the task of fPAD. We propose the Federated Face Presentation Attack Detection (FedPAD) framework to develop the robust and generalized fPAD model in a data privacy preserving way. 

\noindent 2. An experimental setting is defined for the FedPAD framework.  Extensive experiments are carried out to show the effectiveness of the proposed framework. Various issues and insights about federated learning for fPAD are discussed and provided.

\begin{figure*}[!htb]
	
	\begin{center}
		
		\includegraphics[width=0.8\linewidth]{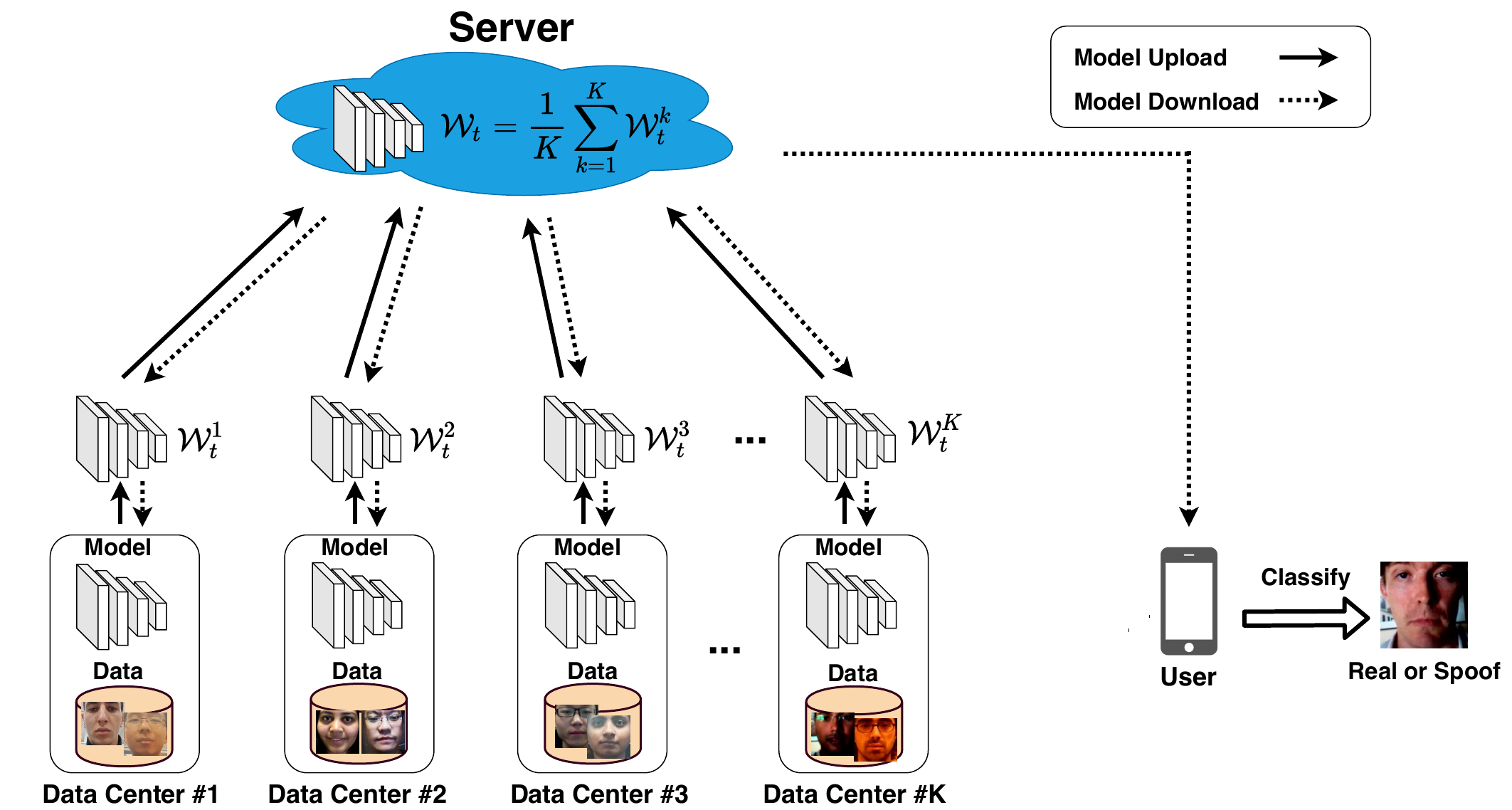}
		
	\end{center}
	
	\caption{Overview of the proposed FedPAD framework. Trough several rounds of communication between \textit{data centers} and \textit{server}, the collaborated trained global fPAD model parameterized by $\mathcal{W}_t$ can be obtained in a data privacy preserving way. Users can download this global model from the server to their device to detect various face presentation attacks.}
	\label{fig:Overview}
\end{figure*}

\section{Related Work}

\subsection{Face Presentation Attack Detection}
Current fPAD methods can be categorized under single-domain and multi-domain approaches. Single-domain approach focuses on extracting discriminative cues between real and spoof samples from a single dataset, which can be further divided into appearance-based methods and temporal-based methods. Appearance-based methods focus on extracting various discriminative appearance cues for detecting face presentation attacks. Multi-scale LBP~\cite{2011IJCBmstexture} and color textures~\cite{2016TIFScolortxt} methods are two texture-based methods that extract various LBP descriptors in various color spaces for the differentiation between real/spoof samples. Image distortion analysis~\cite{2015TIFSida} aims to detect the surface distortions as the discriminative cue.  

On the other hand, temporal-based methods extract different discriminative temporal cues through multiple frames between real and spoof samples. Various dynamic textures are exploited in~\cite{2014EJIVPlbptop,2018TIFSdynamictext,RuiShao2018IJCB} to extract discriminative facial motions. rPPG signals are exploited by Liu \emph{et al.}~\cite{2016ECCVrPPG,2018ECCVrPPG} to capture discriminative heartbeat information from real and spoof videos. ~\cite{2018CVPRauxliary} learns a CNN-RNN model to estimate different face depth and rPPG signals between the real and spoof samples. 

Various fPAD datasets are introduced recently that explore different characteristics and scenarios of face presentation attacks. Multi-domain approach is proposed in order to improve the generalization ability of the fPAD model to unseen attacks. Recent work~\cite{Shao2019CVPR} casts fPAD as a domain generalization problem and proposes a multi-adversarial discriminative deep domain generalization framework to search generalized differentiation cues in a shared and discriminative feature space among multiple fPAD datasets. ~\cite{liu2019deep} treats fPAD as a zero-shot problem and proposes a Deep Tree Network to partition the spoof samples into multiple sub-groups of attacks. ~\cite{Shao_2020_AAAI} addresses fPAD with a meta-learning framework and enables the model learn to generalize well through simulated train-test splits among multiple datasets. These multi-domain approaches have access to data from multiple datasets or multiple spoof sub-groups that enable them to obtain generalized models. In this paper, we study the scenario in which each \textit{data center} contains data from a single domain. Due to data privacy issues, we assume that they do not have access to data from other \textit{data centers}. This paper aims to exploit  multi-domain information in a privacy preserving manner.

\subsection{Federated Learning}
Federated learning is a decentralized machine learning approach that enables multiple local clients to collaboratively learn a global model with the help of a server while preserving data privacy of local clients. Federated averaging (FedAvg)~\cite{mcmahan2016communication}, one of the fundamental frameworks for FL, learns a global model by averaging model parameters from local clients. FedProx~\cite{sahu2018convergence} and Agnostic Federated Learning (AFL)~\cite{mohri2019agnostic} are two variants of FedAvg which aim to address the bias issue of the learned global model towards different clients. These two methods achieve better models  by adding proximal term to the cost functions and optimizing a centralized distribution mixed with client distributions, respectively. This paper focuses on exploiting FedAvg~\cite{mcmahan2016communication} to improve the generalization ability of a fPAD model while securing data privacy of each \textit{data center}.

\section{Proposed Method}

The proposed FedPAD framework is summarized in Fig.~\ref{fig:Overview} and Algorithm~\ref{algorithm}. Suppose that $K$ \textit{data enters} collect their own fPAD datasets designed for different characteristics and scenarios of face presentation attacks. The corresponding collected fPAD datasets are denoted as $\mathcal{D}^{1}, \mathcal{D}^{2},..., \mathcal{D}^{K}$ with data provided with image and label pairs denoted as $x$ and $y$. $y$ are ground-truth with binary class labels (y= 0/1 are the labels of spoof/real samples). Based on the collected fPAD data, each \textit{data center} can train its own fPAD models by iteratively minimizing the cross-entropy loss as follows:

\begin{equation}
	\begin{split}
		\mathcal{L}(\mathcal{W}^k) = \sum\limits_{(x,y)\sim\mathcal{\mathcal{D}}^{k}}y\log\mathcal{F}^k(x)+(1-y)\log(1-\mathcal{F}^k(x)),
	\end{split}
\end{equation} 
where the fPAD model $\mathcal{F}^{k}$ of the $k$-th \textit{data enter} is parameterized by $\mathcal{W}^k$ ($k=1,2,3,...,K$). After optimization with several local epochs via $$\mathcal{W}^k \leftarrow \mathcal{W}^k-\eta \nabla\mathcal{L}(\mathcal{W}^k),$$ each \textit{data enter} can obtain the trained fPAD model with the updated model parameters.

It should be noted that dataset corresponding to each  \textit{data enter} is from a specific input distribution and it only contains  a finite set of known types of spoof attack data. When a  model is trained using this data, it focuses on addressing the characteristics and scenarios of face presentation attacks prevalent in the corresponding dataset. However,  a model trained from a specific \textit{data center} will not generalize well to unseen face presentation attacks. It is well known fact that diverse fPAD training data contributes to a better generalized fPAD model. A straightforward solution is to collect and combine all the data from $K$ data centers denoted as $\mathcal{D} = \{\mathcal{D}^{1} \cup \mathcal{D}^{2}\cup...\cup \mathcal{D}^{K}\} $ to train a fPAD model. It has been shown that domain generalization and meta-learning based fPAD methods can further improve the generalization ability with the above combined multi-domain data $\mathcal{D}$~\cite{Shao2019CVPR,Shao_2020_AAAI}. However, when sharing data between different \textit{data centers} are prohibited due to the privacy issue, this naive solution is not practical.

\begin{algorithm}[t]
	\normalsize 
	\caption{Federated Face Presentation Attack Detection}
	\begin{algorithmic}
		\REQUIRE~~\\
		\textbf{Input:} $K$ Data Centers have $K$ fPAD datasets $\mathcal{D}^{1}, \mathcal{D}^{2},..., \mathcal{D}^{K}$,  \\
		\textbf{Initialization:} $K$ Data Centers have $K$ fPAD models $\mathcal{F}^{1}, \mathcal{F}^{2},..., \mathcal{F}^{K}$ parameterized by $\mathcal{W}_0^1, \mathcal{W}_0^2,..., \mathcal{W}_0^K$. $L$ is the number of local epochs. $\eta$ is the learning rate. $t$ is the federated learning rounds
		
		\textbf{Server aggregates:}
		
		initialize $\mathcal{W}_0$
		\FOR{each round $t$ = 0, 1, 2,... }
		\FOR{each data center $k$ = 1, 2,..., $K$  \textbf{in parallel}}
		\STATE $  {W}_t^k \leftarrow {\rm \textbf{DataCenterUpdate}}(k, \mathcal{W}_t)$
		\ENDFOR
		\STATE $\mathcal{W}_t=\frac{1}{K} \sum\limits_{k=1}^K\mathcal{W}_t^k$
		\STATE 	\textbf{Download} ${W}_t$ to \textbf{Data Centers}
		\ENDFOR
		
		\textbf{Users} \textbf{Download} ${W}_t$
		\STATE
		\STATE
		${\rm \textbf{DataCenterUpdate}}(k, \mathcal{W})$\textbf{:}
		\FOR{each local epoch $i$ = 1, 2,..., $L$  }
		\STATE $\mathcal{L}(\mathcal{W}^k) = \sum\limits_{(x,y)\sim\mathcal{\mathcal{D}}^{k}}y\log\mathcal{F}^k(x)+(1-y)\log(1-\mathcal{F}^k(x))$
		\STATE $\mathcal{W}^k \leftarrow \mathcal{W}^k-\eta \nabla\mathcal{L}(\mathcal{W}^k)$
		\ENDFOR
		
		\textbf{Upload} ${W}^k$ to \textbf{Server}
	\end{algorithmic}
	\label{algorithm}
\end{algorithm}

To circumvent this limitation and enable various data centers to collaboratively train a fPAD model, we propose the FedPAD framework. Instead of accessing private fPAD data of each \textit{data center}, the proposed FedPAD framework introduces a \textit{server} to exploit the fPAD information of all data centers by aggregating the above model updates ($\mathcal{W}^1, \mathcal{W}^2,..., \mathcal{W}^K$) of all data centers. Inspired by the  Federated Averaging~\cite{mcmahan2016communication} algorithm, in the proposed framework, server carries out the aggregation of model updates via calculating the average of updated parameters ($\mathcal{W}^1, \mathcal{W}^2,..., \mathcal{W}^K$) in all data centers as follows:
\begin{equation}
	\mathcal{W}=\frac{1}{K} \sum\limits_{k=1}^K\mathcal{W}^k.
\end{equation}
After the aggregation, server  produces a global fPAD model parameterized by $\mathcal{W}$ that exploits the fPAD information of various data centers without accessing the private fPAD data. We can further extend the above aggregation process into $t$ rounds. Server distributes the  aggregated model $\mathcal{W}$ to every data center as the initial model parameters for the next-round updating of local parameters. Thus, data centers can obtain the $t$-th round updated parameters denoted as ($\mathcal{W}_t^1, \mathcal{W}_t^2,..., \mathcal{W}_t^K$). The $t$-th aggregation in the server can be carried out as follows:
\begin{equation}
	\mathcal{W}_t=\frac{1}{K} \sum\limits_{k=1}^K\mathcal{W}_t^k.
\end{equation}
After  $t$-rounds of communication between data centers and the server, the trained global fPAD model parameterized by $\mathcal{W}_t$ can be obtained without compromising the private data of individual \textit{data centers}. Once training is converged, \textit{users} will download the trained model from the server to their devices to carry out fPAD locally. 

\section{Experiments}
\begin{figure*}[!htb]
	\begin{center}
		\includegraphics[height=2.6cm, width=1\linewidth]{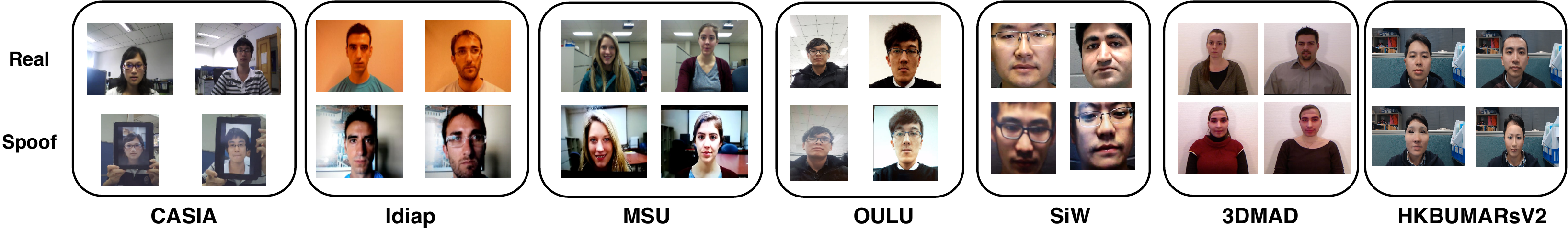}
	\end{center}
	\caption{Sample images corresponding to real and attacked faces from CASIA-MFSD~\cite{2012ICBcasia}, Idiap Replay-Attack~\cite{2012BIOSIGidiap}, MSU-MFSD~\cite{2015TIFSida}, Oulu-NPU~\cite{2017FGoulu}, SiW~\cite{2018CVPRauxliary}, 3DMAD~\cite{Marceltifs3D2014}, and HKBUMARsV2~\cite{2018ECCVrPPG} datasets.}
	\label{fig:datasets}
\end{figure*}

To evaluate the performance of the proposed FedPAD framework, we carry out experiments using five 2D fPAD datasets and two 3D mask fPAD datasets. In this section, we
first describe the datasets and the testing protocol used in our experiments. Then we report various experimental results based on multiple fPAD datasets. Discussions and analysis about the results are carried out to provide various insights about FL for fPAD.

\subsection{Experimental Settings}

\subsubsection{Datasets}

\begin{table}[htb]
	\renewcommand{\arraystretch}{1}
	\centering	
	\scriptsize
	\caption{Comparison of seven experimental datasets.}	
	\begin{tabular}{c|c|c|c|c}
		\hline\hline
		\textbf{Dataset} & \begin{tabular}[c]{@{}c@{}}\textbf{Extra }\\\textbf{ light}\end{tabular} & \begin{tabular}[c]{@{}c@{}}\textbf{Complex}\\ \textbf{background}\end{tabular} & \begin{tabular}[c]{@{}c@{}}\textbf{Attack}\\ \textbf{type}\end{tabular}                                  & \begin{tabular}[c]{@{}c@{}}\textbf{Display} \\ \textbf{devices}\end{tabular}           \\ \hline
		C       & No                                                     & Yes                                                           & \begin{tabular}[c]{@{}c@{}}Printed photo\\ Cut photo\\ Replayed video\end{tabular}     & iPad                                                                 \\ \hline
		I       & Yes                                                    & Yes                                                          & \begin{tabular}[c]{@{}c@{}}Printed photo\\ Display photo\\ Replayed video\end{tabular} & \begin{tabular}[c]{@{}c@{}}iPhone 3GS \\ iPad\end{tabular}           \\ \hline
		M       & No                                                     & Yes                                                          & \begin{tabular}[c]{@{}c@{}}Printed photo\\ Replayed video\end{tabular}                 & \begin{tabular}[c]{@{}c@{}}iPad Air\\ iPhone 5S\end{tabular}         \\ \hline
		O       & Yes                                                    & No                                                           & \begin{tabular}[c]{@{}c@{}}Printed photo\\ Display photo\\ Replayed video\end{tabular} & \begin{tabular}[c]{@{}c@{}}Dell 1905FP\\ Macbook Retina\end{tabular} \\ \hline
		
		S       & Yes                                                    & Yes	                                                           & \begin{tabular}[c]{@{}c@{}}Printed photo\\ Display photo\\ Replayed video\end{tabular} & \begin{tabular}[c]{@{}c@{}}Dell 1905FP\\iPad Pro\\iPhone 7\\Galaxy S8\\ Asus MB168B\end{tabular} \\ \hline
		
		3       & No                                                    & No	                                                           & \begin{tabular}[c]{@{}c@{}}Thatsmyface 3D mask\end{tabular} & \begin{tabular}[c]{@{}c@{}}Kinect \end{tabular} \\ \hline
		H       & Yes                                                    & Yes	                                                           & \begin{tabular}[c]{@{}c@{}}Thatsmyface 3D mask\\REAL-f mask\end{tabular} & \begin{tabular}[c]{@{}c@{}}MV-U3B\end{tabular} \\ \hline\hline
	\end{tabular}
	\label{tab:datasets}
\end{table}

We evaluate our method using the following seven publicly available fPAD datasets which contain print, video replay and 3D mask attacks:\\
1) Oulu-NPU~\cite{2017FGoulu} (O for short)\\
2) CASIA-MFSD~\cite{2012ICBcasia} (C for short)\\ 
3) Idiap Replay-Attack~\cite{2012BIOSIGidiap} (I for short)\\ 
4) MSU-MFSD~\cite{2015TIFSida} (M for short)\\
5) SiW~\cite{2018CVPRauxliary} (S for short)\\
6) 3DMAD~\cite{Marceltifs3D2014} (3 for short)\\
7) HKBUMARsV2~\cite{2018ECCVrPPG} (H for short).\\ 
Table~\ref{tab:datasets} shows the variations in these seven datasets. Some sample images from these datasets are shown in Fig.~\ref{fig:datasets}. From Table~\ref{tab:datasets} and Fig.~\ref{fig:datasets} it can be seen that different fPAD datasets exploit different characteristics and scenarios of face presentation attacks (\textit{i.e.} different attack types, display materials and resolution, illumination, background and so on). Therefore, significant  domain shifts exist among these datasets.  

\subsubsection{Protocol} 
The testing protocol used in the paper is designed to test the generalization ability of fPAD models. Therefore, in each test, performance of a trained model is evaluated against a dataset that it has not been observed during training. In particular, we choose one dataset at a time to emulate the role of \textit{users} and consider all other datasets as \textit{data centers}. Real images and spoof images of \textit{data centers} are used to train a fPAD model. The trained model is tested considering the dataset that emulates the role of \textit{users}. We evaluate the performance of the model by considering how well the model is able to differentiate between real and spoof images belonging to each \textit{user}.

\subsubsection{Evaluation Metrics} 
Half Total Error Rate (HTER)~\cite{2004HTER} (half of the summation of false acceptance rate and false rejection rate), Equal Error Rates (EER) and Area Under Curve (AUC) are used as evaluation metrics in our experiments, which are three most widely-used metrics for the cross-datasets/cross-domain evaluations. Following ~\cite{liu2019deep}, in the absence of a development set,  thresholds required for calculating evaluation metrics are determined based on the data in all \textit{data centers}.

\subsubsection{Implementation Details} 
Our deep network is implemented on the platform of PyTorch. We adopt Resnet-18~\cite{Kaiming_Resnet_CVPR2016} as the structure of fPAD models $\mathcal{F}^{i} (i=1,2,3,...,K)$. The Adam optimizer~\cite{adam} is used for the optimization. The learning rate is set as 1e-2. The batch size is 64 per data center. Local optimization epoch $L$ is set equal to 3.

\subsection{Experimental Results}

\begin{table*}[htb]
	\renewcommand{\arraystretch}{1}
	\centering	
	\caption{Comparison with models trained by data from single data center and various data centers.}
	\begin{tabular}{c|c|c|c|c|c|c|c|c}
		\hline
		\textbf{Methods}                  & \textbf{Data Centers} & \textbf{User} & \textbf{HTER (\%)} & \textbf{EER (\%)} & \textbf{AUC (\%)} & \textbf{Avg. HTER}          & \textbf{Avg. EER}          & \textbf{Avg. AUC}           \\ \hline
		\multirow{12}{*}{\textbf{Single}} & O                     & M             & 41.29              & 37.42             & 67.93             & \multirow{12}{*}{36.43} & \multirow{12}{*}{34.31} & \multirow{12}{*}{70.36} \\
		& C                     & M             & 27.09              & 24.69             & 82.91             &                         &                         &                         \\ 
		& I                     & M             & 49.05              & 20.04             & 85.89             &                         &                         &                         \\ 
		& O                     & C             & 31.33              & 34.73             & 73.19             &                         &                         &                         \\ 
		& M                     & C             & 39.80              & 40.67             & 66.58             &                         &                         &                         \\ 
		& I                     & C             & 49.25              & 47.11             & 55.41             &                         &                         &                         \\ 
		& O                     & I             & 42.21              & 43.05             & 54.16             &                         &                         &                         \\ 
		& C                     & I             & 45.99              & 48.55             & 51.24             &                         &                         &                         \\ 
		& M                     & I             & 48.50              & 33.70             & 66.29             &                         &                         &                         \\ 
		& M                     & O             & 29.80              & 24.12             & 84.86             &                         &                         &                         \\ 
		& C                     & O             & 33.97              & 21.24             & 84.33             &                         &                         &                         \\ 
		& I                     & O             & 46.95              & 35.16             & 71.58             &                         &                         &                         \\ \hline
		\multirow{4}{*}{\textbf{Fused}}    & O\&C\&I               & M             & 34.42              & 23.26             & 81.67             & \multirow{4}{*}{35.75}  & \multirow{4}{*}{31.29}  & \multirow{4}{*}{73.89}  \\ 
		& O\&M\&I               & C             & 38.32              & 38.31             & 67.93            &                         &                         &                         \\ 
		& O\&C\&M               & I             & 42.21              & 41.36             & 59.72             &                         &                         &                         \\ 
		& I\&C\&M               & O             & 28.04              & 22.24             & 86.24             &                         &                         &                         \\ \hline
		\multirow{4}{*}{\textbf{Ours}}    & O\&C\&I               & M             & 19.45              & 17.43             & 90.24             & \multirow{4}{*}{\textbf{32.17}}  & \multirow{4}{*}{\textbf{28.84}}  & \multirow{4}{*}{\textbf{76.51}}  \\ 
		& O\&M\&I               & C             & 42.27              & 36.95             & 70.49             &                         &                         &                         \\ 
		& O\&C\&M               & I             & 32.53              & 26.54             & 73.58             &                         &                         &                         \\ 
		& I\&C\&M               & O             & 34.44              & 34.45             & 71.74             &                         &                         &                         \\ \hline
		\hline		
		\multirow{4}{*}{\begin{tabular}[c]{@{}c@{}}\textbf{All}\\ (Upper Bound)\end{tabular}}     & O\&C\&I               & M             & 21.80              & 17.18             & 90.96             & \multirow{4}{*}{27.26}  & \multirow{4}{*}{25.09}  & \multirow{4}{*}{80.42}  \\ 
		& O\&M\&I               & C             & 29.46              & 31.54             & 76.29             &                         &                         &                         \\ 
		& O\&C\&M               & I             & 30.57              & 25.71             & 72.21             &                         &                         &                         \\ 
		& I\&C\&M               & O             & 27.22              & 25.91             & 82.21             &                         &                         &                         \\ \hline
	\end{tabular}
	\label{tab:singleallours}
\end{table*}

\begin{figure*}[htb]
	\centering
	\begin{minipage}[t]{0.45\linewidth}
		\centering
		\includegraphics[width=1\linewidth]{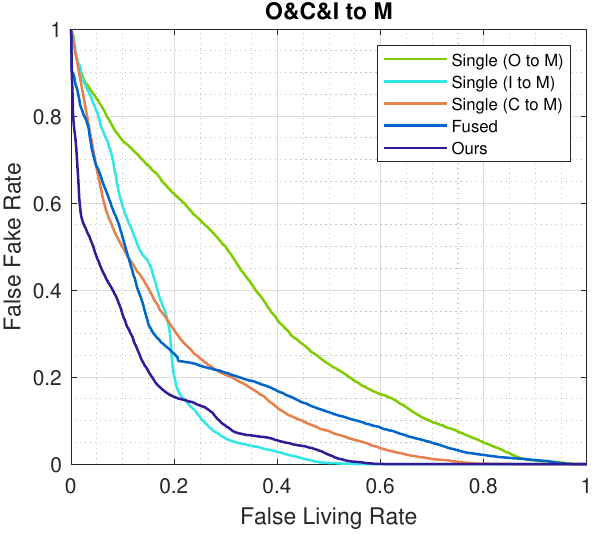}
	\end{minipage}%
	\begin{minipage}[t]{0.45\linewidth}
		\centering
		\includegraphics[width=1\linewidth]{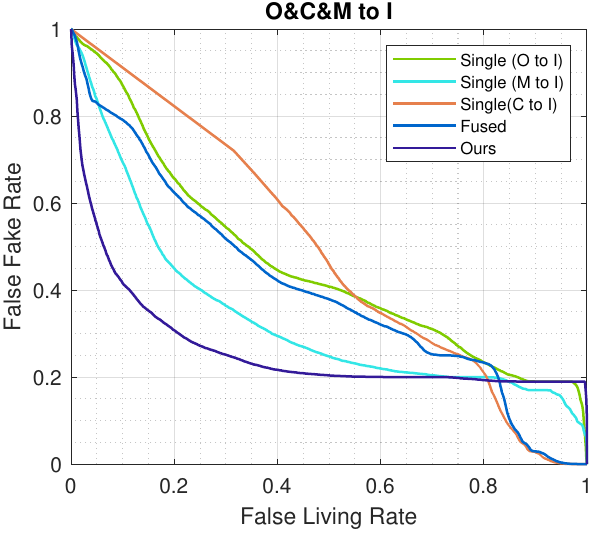}  
	\end{minipage}
	\caption{ROC curves of models trained by data from single data center and various data centers.}
	\label{fig:rocmain}
\end{figure*}

\begin{figure*}[htb]
	\centering
	\begin{minipage}[t]{0.45\linewidth}
		\centering
		\includegraphics[width=1\linewidth]{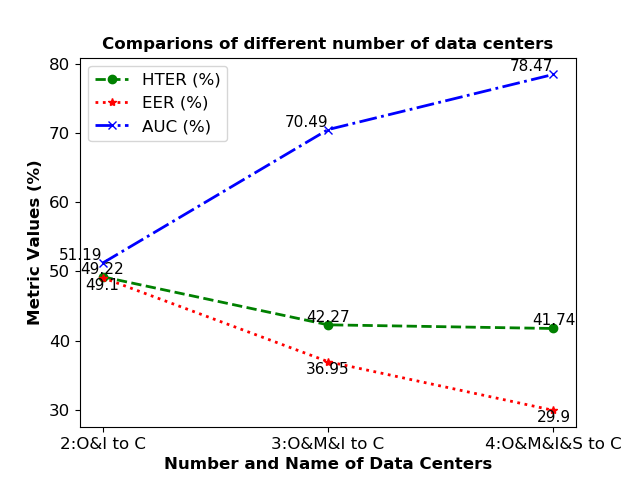}
	\end{minipage}%
	\begin{minipage}[t]{0.45\linewidth}
		\centering
		\includegraphics[width=1\linewidth]{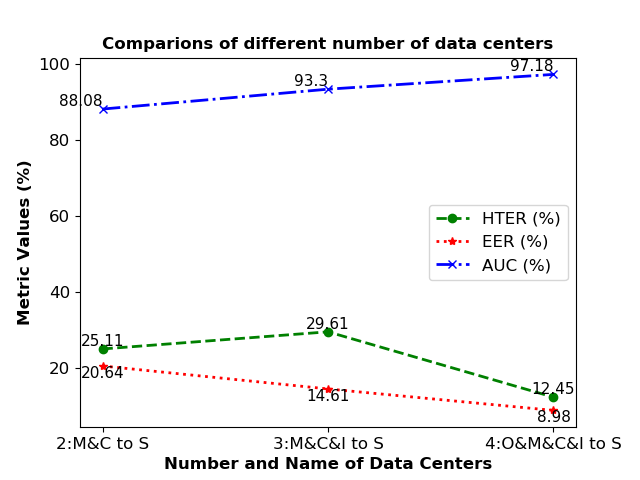}
	\end{minipage}
	\caption{Comparison of different number of data centers.}
	\label{fig:numberofdatacenter}
\end{figure*}

In this section we demonstrate the practicality and generalization ability of the proposed framework in the real-world scenario. We first compare the performance of the proposed framework with models trained with data from a single data center. As mentioned above, due to the limitation of data privacy that exists in the real-world, data cannot be shared between different \textit{data centers}. In this case, \textit{users} will directly obtain a trained model from one of the \textit{data centers}. We report the performance of this baseline in the Table~\ref{tab:singleallours} under the label \textbf{Single}. For different choices of user datasets (from O, C, I, M), we report the performance when the model is trained from the remaining datasets independently. 

Rather than obtaining a trained model from a single data center, it is possible for users to obtain multiple trained models from several data centers and fuse their prediction scores during inference, which is also privacy preserving. In this case, we fuse the prediction scores of the trained model from various data centers by calculating the average. The results of this baseline are shown in Table~\ref{tab:singleallours} denoted as \textbf{Fused}. Please note that it is impossible for users to carry out feature fusion because a classifier cannot be trained based on the fused features without accessing to any real/spoof data during inference in the users. According to  Table~\ref{tab:singleallours}, fusing scores obtained from different data centers improves the fPAD performance on average. However, this would require higher inference time and computation complexity (of order three for the case considered in this experiment).

\begin{table*}
	\renewcommand{\arraystretch}{1}
	\centering	
	\caption{Effect of using different types of spoof attacks}
	\begin{tabular}{c|c|c|c|c|c}
		\hline
		\textbf{Methods}                 & \textbf{Data Centers}  & \textbf{User}    & \textbf{HTER (\%)} & \textbf{EER (\%)} & \textbf{AUC (\%)} \\ \hline
		\multirow{2}{*}{\textbf{Single}} & I (Print)              & M (Print, Video) & 38.82              & 33.63             & 72.46             \\ \cline{2-6} 
		& O (Video)              & M (Print, Video) & 35.76              & 28.55             & 78.86             \\ \hline
		\textbf{Fused}                     & I (Print) \& O (video) & M (Print, Video) & 35.22              & \textbf{25.56}             & 81.54             \\ \hline		
		\textbf{Ours}                    & I (Print) \& O (video) & M (Print, Video) & \textbf{30.51}     & 26.10    & \textbf{84.82}    \\ \hline

	\end{tabular}
	\label{tab:2Dtype}
\end{table*}


On the other hand, \textbf{Ours} shows the results obtained by the proposed FedPAD framework. Table~\ref{tab:singleallours} illustrates that the average values of all evaluation metrics of the proposed framework outperform both baselines. Specifically, we further plot the ROC curves which choose MSU-MFSD (M) and Idiap Replay-Attack (I) datasets as the users as two examples, respectively in Fig~\ref{fig:rocmain}. Fig~\ref{fig:rocmain} also shows that the proposed FedPAD framework performs better than two baselines. This demonstrates that the proposed method is more effective in exploiting fPAD information from multiple data centers. This is because the proposed framework actively combines fPAD information across data centers during training as opposed to the \textbf{fused} baseline. As a result, it is able to generalize better to unseen/novel spoof attacks.  

\begin{table*}[!htb]
	\renewcommand{\arraystretch}{1}
	\centering	
	\caption{Impact of adding data centers with diverse attacks}		
	\begin{tabular}{c|c|c|c|c}
		\hline
		\textbf{Data Centers} & \textbf{User} & \textbf{HTER (\%)} & \textbf{EER (\%)} & \textbf{AUC (\%)} \\ \hline
		O\&C\&I\&M (2D)          & H (3D)        & 47.02              & 18.31             & 85.06             \\ \hline
		O\&C\&I\&M (2D)\&3 (3D) & H (3D)        & \textbf{34.70}              & \textbf{14.20}             & \textbf{92.35}             \\ \hline
	\end{tabular}
	\label{tab:3Dtype}
\end{table*}
Moreover, we further consider the case where a model is trained with data from all available data centers, which is denoted as \textbf{All} in Table~\ref{tab:singleallours}. Note that this baseline violates the assumption of preserving data privacy, and therefore is not a valid comparison for FedPAD. Nevertheless, it indicates the upper bound of performance for the proposed FedPAD framework. From Table~\ref{tab:singleallours}, it can be seen that the proposed FedPAD framework is only $3.9 \%$ worse than the upper bound in terms of AUC.  This shows the proposed framework is able to obtain a privacy persevering fPAD model without sacrificing too much fPAD performance. This result verifies the practicality of the proposed framework.\\

\subsubsection{Comparison of different number of data centers}

\begin{figure}[!htb]
	\begin{center}
		\includegraphics[ width=0.9\linewidth]{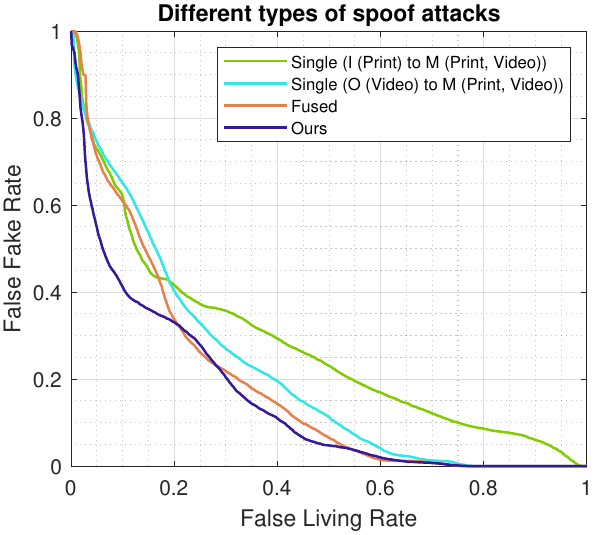}
	\end{center}
	\caption{ROC curves of models trained by different types of 2D spoof attacks.}
	\label{fig:rocprintvideo}
\end{figure}

In this section, we investigate the importance of having more data centers during training. Different data centers exploit different characteristics of face presentation attacks. Therefore, we expect aggregating information from more data centers in the proposed FedPAD framework to produce more robust models with better  generalization. In order to verify this point, we increase the number of data centers in the proposed FedPAD framework and report the results in Fig.~\ref{fig:numberofdatacenter}. The experiments are carried out using five datasets (O, M, I, C, S). In Fig.~\ref{fig:numberofdatacenter} (left), we select the dataset C as the data presented to the user and the remaining datasets as the data centers for training the fPAD model with our FedPAD framework. We increase the number of data centers from 2 to 4 and corresponding \textit{data centers} are shown in the X-axis. Another experiment is carried out with a different combination of the same five datasets and the results are shown in Fig.~\ref{fig:numberofdatacenter} (right). From the curve in Fig.~\ref{fig:numberofdatacenter}, it can be seen that most values of evaluation metrics improve along when the number of data centers increases. This demonstrates that increasing the number of data centers in the proposed FedPAD framework can improve the performance.

\subsubsection{Generalization ability to various 2D spoof attacks}

In reality, due to limited resources, one data center may be only able to collect limited types of 2D attacks. However, various 2D attacks may appear to the users. This section supposes that one data center collects one particular type of 2D attack such as print attack or video-replay attack. As illustrated in Table~\ref{tab:2Dtype}, first, we select real faces and print attacks from dataset I and real faces and video-replay attacks from dataset O to train a fPAD model respectively and evaluate them on dataset M (containing both print attacks and video-replay attacks). In both considered cases as shown in Table~\ref{tab:2Dtype}, the corresponding trained models cannot generalize well to dataset M which contains the additional types of 2D attacks compared to dataset I and O, respectively. This tendency can be alleviated when the prediction scores of two independently trained models on both types of attacks are fused as show in Table~\ref{tab:2Dtype}. Comparatively, FedPAD method obtains a performance gain of $4.71\%$ in HTER and $3.3\%$ in AUC compared to score fusion. We also plot the corresponding ROC curve for this comparison in Fig~\ref{fig:rocprintvideo}, which also demonstrate the superior performance of the proposed method. This experiment demonstrates that carrying out FedPAD framework among data centers with different types of 2D spoof attacks can improve the generalization ability of the trained fPAD model to various 2D spoof attacks.

\subsubsection{Generalization ability to 3D mask attacks }

\begin{figure}[!htb]
	\begin{center}
		\includegraphics[ width=0.9\linewidth]{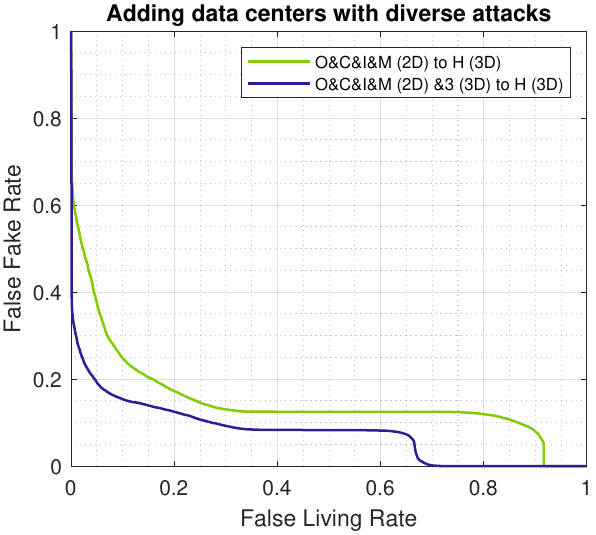}
	\end{center}
	\caption{ROC curves of models trained with added data center of 3D mask attacks.}
	\label{fig:roc2D3D}
\end{figure}

In this section, we investigate the generalization ability of the proposed FedPAD framework to 3D mask attacks. First, a fPAD model is trained with data centers exploiting 2D attacks (data from datasets O, C, I and M). This model is tested  with 3D mask attacks (data from dataset H). Then, we include one more data center containing 3D mask attacks (dataset 3) into our FedPAD framework and retrain our model. Table~\ref{tab:3Dtype} and Fig~\ref{fig:roc2D3D} show that introducing  diversity of data centers  (by including a 3D mask attack) can significantly improve performance in all evaluation metrics. This experiment demonstrates that increasing data centers with 3D mask attacks within the proposed FedPAD framework can improve the generalization ability of the trained model to the novel 3D mask attacks.

\section{Conclusion}
In this paper, we presented FedPAD, a federated learning-based framework targeting application of fPAD with the objective of obtaining generalized fPAD models while preserving data privacy. Through communications between \textit{data centers} and the \textit{server}, a globe fPAD model is obtained by iteratively aggregating the model updates from various \textit{data centers}. Local private data is not accessed during this  process. Extensive experiments are carried out to demonstrate the effectiveness of the proposed framework which provide various insights regarding federated learning for fPAD.

In our experiments, we encountered situations where adding more data centers slightly decreased the performance ( HTER value  of Fig.~\ref{fig:numberofdatacenter} (right) when the data center increases from 2 to 3).  Adding more data centers into our framework not only increases the diversity of fPAD information but also intensifies domain shift among data centers. This domain shift may increase the difficulties in exploiting an optimal fPAD model in the federated learning process. In the future, we will explore an improved federated learning method  that tackles data centers with significant domain shift effectively.

{\small
\bibliographystyle{ieee_fullname}
\bibliography{egbib}

\begin{thebibliography}{10}\itemsep=-1pt

\bibitem{2004HTER}
Samy Bengio and Johnny Mari{\'{e}}thoz.
\newblock A statistical significance test for person authentication.
\newblock In {\em The Speaker and Language Recognition Workshop}, 2004.

\bibitem{2017FGoulu}
Zinelabinde Boulkenafet and et al.
\newblock Oulu-npu: A mobile face presentation attack database with real-world
  variations.
\newblock In {\em FG}, 2017.

\bibitem{2016TIFScolortxt}
Zinelabidine Boulkenafet, Jukka Komulainen, and Abdenour Hadid.
\newblock Face spoofing detection using colour texture analysis.
\newblock In {\em IEEE Trans. Inf. Forens. Security, 11(8): 1818-1830}, 2016.

\bibitem{2012BIOSIGidiap}
Ivana Chingovska, Andr{\'{e}} Anjos, and S{\'{e}}bastien Marcel.
\newblock On the effectiveness of local binary patterns in face anti-spoofing.
\newblock In {\em BIOSIG}, 2012.

\bibitem{Marceltifs3D2014}
N. Erdogmus and S. Marcel.
\newblock Spoofing face recognition with {3D} masks.
\newblock 2014.
\newblock TIFS.

\bibitem{Kaiming_Resnet_CVPR2016}
Kaiming He, Xiangyu Zhang, Shaoqing Ren, and Jian Sun.
\newblock Deep residual learning for image recognition.
\newblock In {\em CVPR}, 2016.

\bibitem{adam}
Diederik~P. Kingma and Jimmy Ba.
\newblock {A}dam: A method for stochastic optimization.
\newblock In {\em arXiv preprint arXiv:1412.6980}, 2014.

\bibitem{li2019federated}
Tian Li, Anit~Kumar Sahu, Ameet Talwalkar, and Virginia Smith.
\newblock Federated learning: Challenges, methods, and future directions.
\newblock In {\em arXiv preprint arXiv:1908.07873}, 2019.

\bibitem{2018ECCVrPPG}
Siqi Liu, Xiangyuan Lan, and Pong~C. Yuen.
\newblock Remote photoplethysmography correspondence feature for {3D} mask face
  presentation attack detection.
\newblock In {\em ECCV}, 2018.

\bibitem{2016ECCVrPPG}
Siqi Liu, Pong~C. Yuen, Shengping Zhang, and Guoying Zhao.
\newblock 3{D} mask face anti-spoofing with remote photoplethysmography.
\newblock In {\em ECCV}, 2016.

\bibitem{2018CVPRauxliary}
Yaojie Liu, Amin Jourabloo, and Xiaoming Liu.
\newblock Learning deep models for face anti-spoofing: Binary or auxiliary
  supervision.
\newblock In {\em CVPR}, 2018.

\bibitem{liu2019deep}
Yaojie Liu, Joel Stehouwer, Amin Jourabloo, and Xiaoming Liu.
\newblock Deep tree learning for zero-shot face anti-spoofing.
\newblock In {\em CVPR}, 2019.

\bibitem{2011IJCBmstexture}
Jukka M{\"{a}}{\"{a}}tt{\"{a}}, Abdenour Hadid, and Matti Pietik{\"{a}}inen.
\newblock Face spoofing detection from single images using micro-texture
  analysis.
\newblock In {\em IJCB}, 2011.

\bibitem{mcmahan2016communication}
H~Brendan McMahan, Eider Moore, Daniel Ramage, Seth Hampson, et~al.
\newblock Communication-efficient learning of deep networks from decentralized
  data.
\newblock In {\em AISTATS}, 2016.

\bibitem{mohri2019agnostic}
Mehryar Mohri, Gary Sivek, and Ananda~Theertha Suresh.
\newblock Agnostic federated learning.
\newblock 2019.

\bibitem{2014EJIVPlbptop}
Tiago~Freitas Pereira and et al.
\newblock Face liveness detection using dynamic texture.
\newblock In {\em EURASIP Journal on Image and Video Processing, (1): 1-15},
  2014.

\bibitem{sahu2018convergence}
Anit~Kumar Sahu, Tian Li, Maziar Sanjabi, Manzil Zaheer, Ameet Talwalkar, and
  Virginia Smith.
\newblock On the convergence of federated optimization in heterogeneous
  networks.
\newblock In {\em arXiv preprint arXiv:1812.06127}, 2018.

\bibitem{Shao2019CVPR}
Rui Shao, Xiangyuan Lan, Jiawei Li, and Pong~C. Yuen.
\newblock Multi-adversarial discriminative deep domain generalization for face
  presentation attack detection.
\newblock In {\em CVPR}, 2019.

\bibitem{RuiShao2018IJCB}
Rui Shao, Xiangyuan Lan, and Pong~C. Yuen.
\newblock Deep convolutional dynamic texture learning with adaptive
  channel-discriminability for 3{D} mask face anti-spoofing.
\newblock In {\em IJCB}, 2017.

\bibitem{2018TIFSdynamictext}
Rui Shao, Xiangyuan Lan, and Pong~C. Yuen.
\newblock Joint discriminative learning of deep dynamic textures for {3D} mask
  face anti-spoofing.
\newblock In {\em IEEE Trans. Inf. Forens. Security, 14(4): 923-938}, 2019.

\bibitem{Shao_2020_AAAI}
Rui Shao, Xiangyuan Lan, and Pong~C. Yuen.
\newblock Regularized fine-grained meta face anti-spoofing.
\newblock In {\em AAAI}, 2020.

\bibitem{smith2017federated}
Virginia Smith, Chao-Kai Chiang, Maziar Sanjabi, and Ameet~S Talwalkar.
\newblock Federated multi-task learning.
\newblock In {\em NIPS}, 2017.

\bibitem{2015TIFSida}
Di Wen, Hu Han, and Anil~K. Jain.
\newblock Face spoof detection with image distortion analysis.
\newblock In {\em IEEE Trans. Inf. Forens. Security, 10(4): 746-761}, 2015.

\bibitem{2012ICBcasia}
Zhiwei Zhang and et al.
\newblock A face antispoofing database with diverse attacks.
\newblock In {\em ICB}, 2012.

\end{thebibliography}
}

\end{document}